# The Representation of Legal Contracts

Aspassia Daskalopulu[1], Marek Sergot[2]

[1]Department of Information Systems and Computing, Brunel University, Uxbridge, Middlesex UB8 3PH.

E-mail: Aspassia.Daskalopulu@brunel.ac.uk.

Tel.: +44- 1895- 274000 extension 2831.

Fax: +44- 1895-251686

[2]Department of Computing, Imperial College of Science, Technology and Medicine, 180 Queen's Gate, London SW7 2BZ.

E-mail: mjs@doc.ic.ac.uk.

Tel: +44-171-594 8218

**Abstract:** The paper outlines ongoing research on logic-based tools for the analysis and representation of legal contracts, of the kind frequently encountered in large-scale engineering projects and complex, long-term trading agreements. We consider both contract formation and contract performance, in each case identifying the representational issues and the prospects for providing automated support tools.

1. **Introduction**

Over the last twenty years or so a growing body of research in Artificial Intelligence has focussed on the representation of legislation and regulations (for a comprehensive discussion see (Sergot, 1991)). The motivation for this has been twofold: on the one hand there have been opportunities for developing advisory systems for legal practitioners; on the other hand the Law is a complex domain in which diverse modes of reasoning are employed, offering ample opportunity to test existing artificial intelligence techniques as well as to develop new ones. A variety of paradigms have been employed for the representation of legal or para-legal expertise with a view to modelling legal data and different modes of reasoning and developing practical applications. Areas that have attracted research interest include information retrieval from large corpora of legal texts and cases, e.g. (Rissland et al., 1995; Hafner, 1987), interpretation of legal text, e.g. (Allen et al., 1993), argumentation, e.g. (Prakken, 1997; Sartor, 1994), and legislative drafting, e.g. (Allen, 1982). Current research trends include the creation of legal ontologies, that is explicit conceptualisations of the legal domain (see for example (Visser et al., 1997)).

The idea of applying similar techniques to the representation of legal contracts has come up from time to time, given that contracts serve a function similar to that of legislation: they are meant to regulate the actions of (usually) two parties while they interact (usually in a professional context). The topic however has not been explored in depth. Some authors have even seemed to suggest that the development of electronic tools to support contractual activity is uninteresting, either because the domain of investigation—contractual content—is comparatively trivial, or because the tasks associated with the domain—contractual activity—are straightforward and do not require automation. The ESPRIT-funded ALDUS project (ALDUS, 1992) investigated the potential for developing systems to assist with the drafting of contracts. It concluded that there were no real opportunities for developing economically useful tools. Our view, however, is that such projects have looked at the wrong kind of contracts. ALDUS concerned itself almost exclusively with the Sale of Goods, where contracts do tend to be very simple. But not all contracts are as simple as that. In other areas both contractual content and contractual activity can be extremely complex, and automated support can be time-saving and cost-effective. The development of appropriate tools is challenging: knowledge elicitation and representation require the integration of many paradigms from diverse areas of Artificial Intelligence and confront a number of fundamental representational problems.

This paper reports on research that aims to develop logic-based tools for the analysis and representation of legal contracts. Section 2 presents the areas of contractual activity where automated support might be sought and sets the context for the ensuing discussion. Section 3 discusses the representation requirements of different kinds of tools that seek to support contract formation and contract performance. Section 4 presents a representation scheme that has been developed for legal contract assembly.

## 2. Contractual activity

In common usage the term 'contract' refers both to a legally binding *agreement* between (usually) two parties and to the *document* that records such an agreement, if it is put in writing. In this paper the terms *agreement* and *document* are used when such a distinction needs to be made explicit, and they should be understood as referring to *contractual* agreement and document respectively.

The common perception of contractual activity is that it can roughly be regarded as comprising two phases: Contract *formation*, where the parties involved specify their requirements of each other, negotiate on the various aspects of the exchange which will take place and come to some agreement. And contract *performance*, where the agreement is in force and the business exchange between the parties actually takes place. Consequently there are two broad classes of electronic tools that one could consider, one for each phase of contractual activity.

Contract formation tools include those that
(i) determine whether a given agreement is legally binding (whether a legally valid offer and acceptance exist);
(ii) enable parties to specify their requirements and check whether these are compatible or suggest adjustments in order to make them so (one could choose to call these *negotiation* tools); and, in the case of written agreements
(iii) assist drafters in putting the final product of the negotiation, the document, in written form (one could choose to call these *drafting* tools).

Contract performance tools are those that, given a specific agreement

(i) advise parties about their behaviour during the business exchange, reminding them of what needs to be done and when (one could call these *diary* tools);

(ii) monitor the parties' compliance with the agreement and, in the case of violations, suggest available remedies or advise on the possible consequences.

Not all of these tools are useful for all kinds of contracts. According to legal theorists a contractual situation arises when (usually) two parties enter voluntarily into an agreement, assuming obligations towards each other, for the purpose of exchanging some product or service for a (usually) financial reward (cf. Atiyah, 1989; Stone 1994). Hence contractual situations can be identified in business exchanges ranging from the relatively straightforward (the purchase of a ticket for a bus journey, a simple Sale of Goods, standardised tenancy agreements) to the complex (the establishment of a long-term trading agreement between organisations or a complex trading procedure involving third parties). For contracts at the simple end of this scale, electronic support is likely to be unwanted. Where contracts are based on standard terms and conditions parties form and execute them without any apparent difficulty, when it comes to monitoring compliance the question is often whether goods were delivered on time and whether the required payment was made. So projects such as ALDUS are right in concluding that contractual activity is hardly in need of electronic support—*insofar as* one focuses on contracts at the simplest end of the scale.

The research reported in this paper has been addressing the representation of contracts at the other end of the scale, with particular attention to contracts that govern long-term exchanges between parties. Such agreements are frequently encountered in engineering. Most of the sample documents used as experimental material were provided through a collaborative project supported by British Gas. They concern the supply of natural gas from hydrocarbon field owners. The sample documents that were examined run to 200-300 pages each and are often accompanied by drawings, and various technical appendices. These are not one-off exchanges. The 'life span' of an agreement can be up to 20-25 years with a review every five years or so. Consequently the contracts cover a large number of aspects. Some are typical of contracts in general—such as a specification of the period of the agreement, delivery quantities, prices, billing and payment arrangements, and so on. Some are *particular* to engineering contracts—such as provision of technical services, arrangements for technical reviews, the appointment of experts, the arbitration of disputes, the resolution of *force majeure* claims, insurance arrangements, warranties, indemnities, and so on. Some are specific to the particular kind of engineering—extraction of natural gas from a hydrocarbon field cannot be turned on and off at will, so there are many complex provisions dealing with shutdown procedures, adjustments for over- and under-delivery, monitoring of quality, …

The size and complexity of contracts in these and other areas highlights the need for electronic support in all aspects of contractual activity. As regards contract formation, the process of negotiating and establishing a new agreement is long and requires careful preparation. It is typically undertaken by a team rather than one individual (and this raises problems of co-authorship and co-ordination). The associated costs often account for a significant proportion of the cost of the project as a whole. Though there is no formal requirement in English Law for an agreement to be put in writing in order for it to be legally binding (cf. Atiyah, 1989; Stone, 1994), complex business and engineering agreements *are* recorded in written form. It is essential to provide an agreed point of reference, especially where there are large amounts of detail or where the agreement is to remain in force for a considerable length of time, during which adjustments need to be made (for example to pricing information because of inflation or to other details if some of the circumstances of the organisations involved change). During the negotiation several drafts of the contractual documents are produced. This is because contractual documents are sensitive and an omission or a mistake might have significant financial and legal consequences; moreover as the agreement covers a multitude of interrelated aspects, changes in some parts often propagate to other parts of the document. There is a need both for negotiation tools and for drafting tools.

It is often the case that any large organisation will have a large number of related agreements ongoing at any time. The associated documents are frequently consulted—some parts of them even on a daily basis—both by experienced and junior staff with varying degrees of understanding of the legal contents of the agreement. During the performance of the contract, *force majeure* procedures or even litigation are not unusual and the associated costs are very high. Tools that enable retrieval of contract content (both values for certain parameters and the legal implications of the agreement for the parties involved) are therefore desirable.

The next section considers the representation requirements of tools that support contract formation and contract performance.

## 3. Contractual Content and Representation Requirements

The development of any advisory system for some domain requires a representation of the domain and problem-solving expertise associated with it. This raises problems of knowledge elicitation (*what* information to represent and *where* to obtain it from) and knowledge representation and application (*how* to represent and reason with such information). Developing contract formation and contract performance tools necessitates the elicitation and representation of contractual content. The question therefore arises of what *is* contractual content. Is it the content of *agreements* and/or the content of *documents* that is being sought? One might expect that in the case of written agreements the relevant documents record accurately all that was agreed between the parties. Indeed the Law of Contract (in England and in most other jurisdictions) states that if the contents of a given written agreement need to be established by a court of law, the *parole evidence* rule applies, that is, the courts generally hold that the contractual document contains *all* that the parties agreed and only that (cf. Atiyah, 1989; Stone, 1994). However, the Law of Contract also states that what is agreed includes *express* terms—those which are explicitly recorded in the contractual document—and *implied* terms—those that one might expect to find in all contracts of a particular kind, especially where a legislative Act exists, and with which the parties agree to comply, without recording them explicitly (an example is the Sale of Goods Act which contains provisions regarding the *fitness for purpose* of the goods sold). For some purposes it might therefore be sufficient to represent the contents of contractual documents and only those, whereas for other purposes a more extensive representation is required integrating specific domain knowledge about the nature of the business exchange and relevant legislation.

What follows presents informally the functions that provisions in contractual documents can have and considers the kinds of tools outlined in the previous section. The following discussion is based on engineering contracts from our collaborative project but they are representative of contracts in general.

Our sample contracts indicated that contractual provisions play the following roles, amongst others:

(i) They *define* various terms, that is they fix the meaning for some particular terms in the context of the agreement. The term "Day" for example can be defined to mean some period of time not necessarily comprising 24 hours.

(ii) They *prescribe* certain behaviour for the parties, usually under certain circumstances, or during particular periods for the duration of the contract.

(iii) They specify *procedures* that need to be followed by the parties when certain states of affairs are to be established, such as the appointment of an arbitrator to settle a dispute, the process of a financial claim, the change of delivery times and quantities, the early termination of the agreement and so on.

(iv) They contain *formulae* that are used to calculate values for various parameters, such as the price of goods during particular periods or adjustments to prices or quantities.

(v) They specify conditions, under which other provisions apply. These are sometimes referred to as *secondary* provisions.

The list above is not intended to be a formal classification of contractual clauses, nor is it exhaustive. A particular contract clause may be of more than one of the aforementioned kinds; a provision may prescribe an obligation and at the same time define a term (and likewise for the other kinds of functions).  The purpose of the list is to show the different forms and functions that may need to be captured when document content is represented.

A contract can be regarded as a collection of different conceptual models that are interrelated, and various paradigms can be employed to represent them .  At one level of abstraction, a contract is an organized collection of concepts .  At another level, a contract is a collection of obligations, permissions, entitlements, powers and so on (Jones et al., 1992, 1993, 1996; Jones, 1990).  These are notions that have been studied extensively in legal theory (Hohfeld, 1913; Kanger, 1985; Kanger et al., 1966, Lindahl, 1977; and many others).  At a third level a contract can be regarded as a collection of procedures (protocols) that specify its operational aspects (how the business exchange is to be conducted in practice).  These have temporal and action aspects that are at the core of much current research in Artificial Intelligence, Computer Science and Philosophical Logic.  And from another standpoint still, a contract may be represented as a collection of parameters (the parties, the product in question, the price, the delivery quantities, the delivery date and so on).  Contractual activities are not all concerned with all aspects of a contract.  Each focuses on some particular parts.  Alternative views of contracts need to be represented and sometimes integrated into a single system to support a variety of functions.  However, as mentioned earlier, information that is not contained in contractual documents is also required to support some aspects of contractual activity.  In what follows we expand on this with reference to the tools outlined earlier.

Those tools that are intended to identify legally binding agreements need to establish whether a valid offer and acceptance exists.  The Law (whether in England or elsewhere) specifies explicitly circumstances under which an agreement would *not* be legally binding (for example when it is formed under *duress* or when it involves unlawful activity, or when one of the parties is a minor) (cf. Atiyah, 1989). Such tools need therefore to operate on a representation of the notions of "offer", "acceptance", "validity", "duress", "minor" and "unlawful activity". This suggests that the general problem of identifying legally binding agreements is one of *classification*. Fundamental concepts must be represented and organized in a way that enables other concepts to be defined in terms of them or as instances of them. This is the kind of representation task that accounts for the much of Artificial Intelligence and Law research over the past 20 years. Logic programs have been employed for this purpose (Sergot et al., 1986) as well as a variety of other formalisms (for an overview see (Sergot, 1991)). Gardner's work (1987) for example concentrated on offer/acceptance representation using augmented transition networks and a special rule-based language. The central problem in the representation of *classification norms* is the treatment of *open-textured* concepts, that is concepts whose meaning is not provided by a legislative definition but rather through example and decisions of courts of law (cf. Sanders, 1991; Bench-Capon et al., 1988; Gardner, 1987; McCarty, 1980).

Tools that are intended to support contract negotiation, on the other hand, have very different representation requirements. One way to view contracts is as a collection of separate but interrelated sub-agreements. The parties involved have a common goal, to realise the business exchange, to co-operate, but each wants this to happen under the "best" terms for them. What makes a particular arrangement good for a party is relevant to how it affects their broader business goals. Often the goals of the two parties are not mutually satisfiable as they stand, and revision (some mutual compromise) is required. A negotiation tool would therefore be useful if it allowed parties to specify their goals and determined whether these are satisfiable, or would be satisfiable, if certain terms were agreed; if resolution of some conflict were required then it would be useful for the tool to indicate alternative terms (that entail change in the original set of goals). Obviously in its full generality this is a huge problem raising a whole range of issues to which various techniques could be applied. One promising approach is to take an argumentation view of negotiation. Argumentation has been researched intensively in recent years. Prakken (1997) provides an excellent account of legal argumentation in particular and also an overview of recent advances in argumentation generally. As negotiating parties argue for their own interests, their success in getting the "best" for them relies on how *persuasive* their arguments are. A representation scheme that can model persuasive argument might prove useful in this context (for a discussion see (Reed, 1997) in this issue).

Drafting tools raise similar concerns as negotiation tools, and this is no coincidence. The drafters' quest is for "well-formed" documents, and good form entails, among other things, requirements of consistency and completeness—that contractual provisions are not contradictory and that they cover all cases that they are intended to cover. There are theoretical proposals for defining consistency and completeness for a set of *norms* (Alchourron et al., 1971) but application of these methods is not a practical possibility. It would require an exhaustive generation of all the possible factual circumstances (the 'Universe of Cases' in Alchourron's and Bulygin's terms) which is quite unrealistic except in some very special cases. The methods can be applied if we restrict attention to some very narrow, very specific part of the contract. For example one problem which we studied concerned the formulation of a complex set of pricing provisions where the outcome is determined by the occurrence of certain combinations of events and the times at which they occur. Here it was possible to generate all possible combinations of events and times, to establish that they had all been accounted for and that the provisions were not contradictory.

There is also the question of designing a set of norms, that is deciding what obligations, permission, entitlements, rights and so on should be included in the contract. Some automated support for this question can also be provided. Sergot (1997, 1998) presents a generalised and extended version of the Kanger-Lindahl theory of normative positions (Lindahl, 1977). This a theory which attempts to apply a combination of deontic logic (the logic of obligation and permission) and the logic of action/agency to the formalisation of what Hohfeld (1913) called 'the fundamental legal conceptions': duty, right, and other complex normative concepts. The generalised theory includes automated inference methods which have been implemented in computer programs intended to facilitate application of the theory to the analysis of practical problems, either for the purpose of interpretation and disambiguation of legal texts, or in the design and specification of a new set of norms. The objective is to clarify and expand an incomplete and imprecise statement of requirements into a precise formal specification at some desired level of detail. The role of the system is to guide this process, ensuring overall consistency and identifying any possibilities that remain to be explored.

It is difficult to demarcate precisely between the design of the agreement (and hence the object of negotiation tools) and the design of the document (and hence the object of drafting tools). Though the document records the agreement, it is often the case that what is being negotiated *is* the document itself, for example in terms of the text that expresses provisions. As contractual documents are taken to contain all that certain parties agreed by Law, it is important for the text to express as closely as possible what the parties' intended. Section 4 considers the drafting problem in greater detail and presents an approach that concentrates the design of documents.

Contract performance tools aim to advise parties on the effects of individual provisions, once an agreement is in force, to assist in planning the daily business exchange (in terms of what actions need to be performed and when) and to monitor the parties' compliance with the contract. We want to be able to extract parameter values, formulae for pricing or delivery times and detailed procedures as they apply in changing circumstances. We might want to monitor the parties' *compliance* with provisions of the contract. This is not a straightforward matter. It touches on fundamental problems in the field of deontic logic—contrary to duty obligations ( for example Prakken et al., 1996; 1997), the interplay between time and obligation (for example van Eck, 1982), the proper treatment of legal competence or 'power' . 'Power' refers to the ability of a party to create legal relations: for instance, in cases of breach, to vary the terms of an agreement, to terminate the contract, or to take other prescribed remedial action, usually by means of specific pre-agreed procedures (Jones et al., 1996). It is an open question whether such issues need to be resolved before practical applications can be attempted.

## 4. Contract Drafting

This section presents a representation scheme for contractual documents intended to support contract drafting. The proposed representation has been adopted in a prototype system whose implementation details are documented in (Daskalopulu et al., 1995). As mentioned in the previous section, it is hard to separate questions about agreement design from questions about document design and each aspect of the problem has different representation requirements. However there are some issues that are pertinent only to the design of documents and it is on those that this section concentrates.

The approach taken here views contract drafting as conducted at two levels:

(i) at the macro-level, drafters need to establish the coarse structure of the documents, and their overall coherence;

(ii) at the micro-level, the emphasis is on formulating individual provisions in detail, while maintaining consistency and completeness.

Contract drafting at the micro-level is the area where questions relevant to the design of agreements have been touched on already. Micro-drafting of the document concerns the choice of specific words to express provisions. There are proposals in the literature that address such concerns. For example, Layman Allen has been arguing for the use of symbolic logic in legislative drafting since 1957 (Allen, 1957; Allen et al., 1993). Some of these proposals have been applied in practice to the drafting of legislation notably by legislatures of some of the United States. We have not explored the use of these techniques and the associated support software for the micro-drafting of contracts. The tasks are however identical so there seems to be no obstacle to apply them in this way.

We view contract drafting at the macro-level as a form of computer-aided design, where the drafter uses basic blocks of text to construct a document in much the same way that a graphics designer uses basic geometric shapes to construct a picture. To emphasise the use of pre-constructed building blocks it might be more appropriate to call such a process "assembly" rather than "drafting". The idea is not novel. A similar view has been expressed by other researchers (cf. Gordon, 1989; Fiedler, 1985; Lauritsen, 1992) who nevertheless did not develop it into practical applications.

In engineering, for example, it is standard practice for contracts to be drafted using model-forms issued by the relevant professional bodies (the Institution of Electrical Engineers, for example, has been producing such model-forms since 1903). These forms (e.g. (IEE, 1991)) are typically accompanied by a detailed commentary which explains the role of each individual provision in the document, its history and its overall effect. Where model-forms are not available a pre-existing contract of a similar type is frequently used.

In our terms, model-forms or pre-existing documents provide the starting point for developing *generic documents* These are descriptions of classes of documents. Drafting a new contract corresponds to creating a new document instance from this class. Apart from changes in specific data values—or 'parameters'—many of the provisions are acceptable in the standard form. But there are also sub-units or passages of the document which do not suit the circumstances at hand and which require some modification. Such modification may range from being comparatively minor, such as a change of a few words, to being drastic, where whole passages are completely re-written. When drafters modify a given portion of a document, they create in effect different *versions* of that portion of the document. These are stored in the generic document and so become available as models for drafters in the future.

For example Section 4-1 ('Precedence of Documents') in the model form contract (IEE, 1988) reads:

> Unless otherwise provided in the Contract the Conditions as amended by the Letter of Acceptance shall prevail over any other document forming part of the Contract and in the case of conflict between the General Conditions the Special Conditions shall prevail. Subject thereto the Specification shall prevail over any other document forming part of the Contract.

But in the actual contract, a different text had been included in this section:

> The documents forming the Contract are to be taken as mutually explanatory of one another and in the case of ambiguities or discrepancies the same shall be explained and adjusted by the Engineer who shall thereupon issue to the Contractor appropriate instructions in writing.

In another example, Section 14-6 ('Rate of Progress') of the model-form contract (IEE, 1988), which originally reads:

> The Engineer <u>shall</u> notify the Contractor if the Engineer <u>decides</u> that the rate of progress of the Works or of any Section is too slow to meet the Time for Completion and that this is not due to a circumstance for which the Contractor is entitled to an extension of time under Sub-Clause 33-1. (emphasis added) had been modified to replace the occurrences of 'shall' and 'decides' by 'may' and 'considers' respectively. The point is that in neither case is there any indication as to why the modified version had been preferred over the original wording. In supporting CAD-like contract drafting it is important to record reasons for such modifications so that in subsequent drafting situations, drafters can make informed choices about which versions of existing provisions to select to create a new document instance, and about which provisions to formulate in detail (at the micro-level, creating their own version) because none of the existing ones are appropriate.

A representation scheme is therefore required for generic documents (which "grow" over time, as new versions of their provisions are created) and document instances. Generic documents are represented as collections of

(i)      assertions (in a logic database) that reflect the *structural* arrangement of their contents (a document comprises parts, which in turn comprise sections, and those comprise individual provisions which can be further analysed in terms of their constituent sentences and so on) rather than the text itself; and

(ii)     constraints that govern the interrelationships of document sub-components.

Document instances are represented as structured terms that comprise identifiers for the versions of the provisions that the drafter selects or creates. Such identifiers effectively point to the actual text files for those provisions which are held separately. The text of the document instance can be reconstructed in its entirety by a simple program that retrieves the appropriate fragments of text, instantiates any parameters with specific values and collates the fragments into the final document.

A natural question for such representations is what document unit to take as the basic building block. For the engineering contracts in our implementation the section seems to be the most appropriate unit generally but a feature of the scheme is that one does not need to commit to any particular choice of unit. In fact different parts of the same document can be represented at different levels of detail.

The assembly process is guided by constraints. These, as stated earlier, reflect the ways in which document sub-components are interrelated and compliance with them is a necessary requirement for the coherence of the document instance. Such constraints reflect:

(i)  Textual dependency between document sub-components, for example when one document sub-unit contains a cross-reference to another. If the first sub-unit is included in the document instance then the referenced unit must also be included to maintain coherence.

(ii) "Semantic" or "pragmatic" dependency between document sub-components. For example where the drafter includes a provision about third party agreements in the document instance, then a provision about third party liability must be included. The presence of a document sub-unit may (a) necessitate the presence of another document sub-unit, or (b) preclude it. A third possibility is "exclusive or" where exactly one of a number of alternative sub-units must be included in the document instance.

(iii) General (common sense) domain requirements (the two contracting parties must be distinct, the document must specify the date on which the agreement comes into force, this must be later than the date on which it was drafted, ...).

Constraints are represented as part of the generic document and so inherited by all document instances. Constraint checking can be done at the drafter's request or automatically in an incremental fashion as the assembly process takes place. The drafter can choose to be notified about the result of checking by messages detailing pathological features or to have compliance with constraints automatically enforced. Further details of the system are provided in (but cf. Daskalopulu et al., 1995).

## 5. Concluding Remarks

This paper discussed the development of tools to assist in various aspects of contractual activity, both in relation to contract formation and in relation to contract performance. Such tools require different information to be represented: domain knowledge, the relevant legislation (the Law of Contract), the parties' beliefs, preferences and business goals. The common ground for all tools is a representation of contractual content. This term lends itself to ambiguity. Contractual content can refer to the content of the agreement between parties and/or to the content of the document that records the agreement. Though contractual documents are supposed to reflect the corresponding agreement accurately, it is not unusual for them to be vague about aspects that one would expect to be specified explicitly, such as sanctions in the case of violations.

Some aspects of contractual activity can be supported effectively using contractual documents as the primary sources of information. We described a representation scheme that supports contract drafting as a CAD-like assembly process subject to constraints. The approach does not represent the detailed text but rather the structure and the interrelationships between constituent parts of contractual documents. It offers practical support wherever contract drafting is done on the basis of previous examples or model forms. We refer to this as drafting at the macro-level. Contract drafting at the micro-level, that is the formulation of detailed provisions presents a real challenge. A number of tools can be provided for certain tasks and we mentioned some, but in general it is difficult to demarcate the design of the document from the design of the agreement. There is an unlimited range of possible tools that could be deployed to support agreement negotiation and design.

Some projects in the past have concluded that there is no significant demand for automated tools that deal with legal contracts. We argue that this is not so—if we consider the kind of contracts that are typically encountered in engineering projects, long-term business agreements and complex third party trading arrangements.

## 6. References

Alchourrón, C. E. & Bulygin, E. (1971). Normative Systems. Springer-Verlag, New York.

ALDUS (1992). The ALDUS project: Artificial Legal Draftsman for Use in Sales, ESPRIT Commission.

Allen, L. E. (1957). Symbolic Logic: A Razor-Edged Tool for Drafting and Interpreting Legal Documents. *The Yale Law Journal*. **66.** 833-879.

Allen, L. E. (1982). Towards a Normalized Language to Clarify the Structure of Legal Discourse. In Martino (ed.) *Deontic Logic, Computational Linguistics and Legal Information Systems*, North-Holland.

Allen, L. E. & Saxon, C. S. (1993). A-Hohfeld: A Language for Robust Structural Representation of Knowledge in the Legal Domain to Build Interpretation-Assistance Expert Systems. In Meyer & Wieringa (eds) *Deontic Logic in Computer Science: Normative Systems Specification*, John Wiley & Sons.

Atiyah, P. S. (1989). An introduction to the Law of Contract.. Clarendon Press (4th edition), Oxford.

Bench-Capon, T. J. M. & Sergot, M. J. (1988). Towards a rule-based representation of open texture in law. In *Computer Power and Legal Language*, Quorum Books, New York.

Daskalopulu, A. & Sergot, M. J. (1995). A Constraint-Driven System for Contract Assembly. In *Proceedings of the Fifth International Conference on Artificial Intelligence and Law*, University of Maryland, College Park, ACM Press. 62-69.

deKleer, J. (1986). An assumption based TMS. *Artificial Intelligence*. **28.** 127-162.

van Eck, J. A. (1982). A System of Temporally Relative Modal and Deontic Predicate Logic and its Philosophical Applications. *Logique et Analyse*. **100**. 249-381.

Fiedler, H. (1985). Expert Systems as a Tool for Drafting Legal Decisions. In Martino (ed.) *Proceedings of the Second International Conference on Logic, Informatics and Law*, Florence. 265-274.

Gardner, A. (1987). An Artificial Intelligence Approach to Legal Reasoning. MIT Press, Cambridge Massachusetts.

Gordon, T. F. (1992). A Theory Construction Approach to Legal Document Assembly. In Martino (ed.) *Expert Systems in Law*, Elsevier Publishers B.V.

Hafner, C. D. (1987). Conceptual Organization of Case Law Knowledge Bases. In *Proceedings of the First International Conference on Artificial Intelligence and Law,* Boston Massachussetts*,* ACM Press.

Hohfeld, W. N. (1913). Some Fundamental Legal Conceptions as Applied in Judicial Reasoning. *Yale Law Journal*. **23.**

IEE (1988). Model Form of General Conditions of Contracts: Home or Overseas Contracts with Erection (MF/1). The Institution of Electrical Engineers, London.

IEE (1991). Model Form of General Conditions of Contract: Home or Overseas Contracts for the Supply of Electrical of Mechanical Plant (MF/2). The Institution of Electrical Engineers, London.

Jones, A. J. I. (1990). Deontic Knowledge and Legal Knowledge Representation. *Ratio Juris*. **3:2**. 237-244.

Jones, A. J. I. & Sergot, M. J. (1992). Deontic Logic in the Representation of Law: Towards a Methodology. *Artificial Intelligence and Law*. **1:1**. 45-64.

Jones, A. J. I. & Sergot, M. J. (1993). On the Characterisation of Law and Computer Systems: The Normative Systems Perspective. In Meyer & Wieringa (eds) *Deontic Logic in Computer Science: Normative Systems Specification*, John Wiley & Sons.

Jones, A. J. I. & Sergot, M. J. (1996). A Formal Characterisation of Institutionalized Power. *Journal of the Interest Group in Pure and Applied Logics.* **4:3**. 429-445.

Also in: Valdés, Krawietz, von Wright, Zimmerling (eds), *Normative Systems in Legal and Moral Theory*, Duncker & Humblot, Berlin. 1997.

Kanger, S. (1985). On Realization of Human Rights. In Holmstrom & Jones (eds) *Action, Logic and Social Theory*. Acta Philosophica Fennica, **38.**

Kanger, S. & Kanger, H. (1966). Rights and Parliamentarism. *Theoria*. **32.** 85-115.

Lauritsen, M. (1992). Technology Report: Building Legal Practice Systems with Today's Commercial Authoring Tools. *Artificial Intelligence and Law*. **1:1**. 87-102.

Lindhal, L. (1977). Position and Change—A Study in Law and Logic. *Synthese Library*, **112**, D. Reidel, Dordrecht.

McCarty, L. T. (1980). The TAXMAN Project: Towards a Cognitive Theory of Legal Argument. In Niblett (ed.) *Computer Science and Law*, Cambridge University Press, New York.

Prakken, H. (1997). Logical Tools for Modelling Legal Argument: A Study in Defeasible Reasoning in Law. Kluwer, Dordrecht.

Prakken, H. & Sergot, M. J. (1996). Contrary-to-duty Obligations. *Studia Logica*. **57:(1/2).** 91-115.

Prakken, H. & Sergot, M. J. (1997). Dyadic Deontic Logic and Contrary-to-Duty Obligations. In Nute (ed.) *Defeasible Deontic Logic*, Kluwer, The Netherlands.

Reed, C. A. (1997). Representing and Applying Knowledge for Argumentation in a Social Context. *AI and Society*. This issue.

Rissland, E. L. & Daniels, J. L. (1995). A Hybrid CBR-IR Approach to Legal Information Retrieval. In *Proceedings of the Fifth International Conference on Artificial Intelligence and Law*, University of Maryland, College Park, ACM Press.

Sanders, K. E. (1991). Representing and reasoning about open-textured predicates. In *Proceedings of the Third International Conference on Artificial Intelligence and Law,* Oxford, ACM Press.

Sartor, G. (1994). A formal model of legal argumentation. *Ratio Juris.* **7.** 212-226.

Sergot, M. J. (1991). The Representation of Law in Computer Programs. In Bench-Capon (ed.) *Knowledge-Based Systems and Legal Applications*, Academic Press.

Sergot, M. J. (1997). A Computational Theory of Normative Positions. Technical Report, Department of Computing, Imperial College London. Submitted for publication.

Sergot, M. J. (1998). A Method for Automating the Analysis of Normative Positions. In *Proceedings of the Fourth International Workshop on Deontic Logic in Computer Science*, Bologna, Italy (January 1998).

Sergot, M. J., Sadri, F., Kowalski, R. A., Kriwaczek, F., Hammond, P.& Cory, H. T. (1986). The British Nationality Act as a Logic Program. *Communications of the ACM*. **29:5.** 370-386.

Stone, R. (1994). *Contract Law*. Cavendish Publishing Ltd, London.

Visser, P. & Bench-Capon, T. J. M. & van den Herik, J. (1997). A Method for Conceptualising Legal Domains: An Example from the Dutch Unemployment Benefits Act. *Artificial Intelligence and Law*. **5.** 207-242.